\documentclass[final,3p,times]{elsarticle}
\usepackage{graphicx}
\usepackage{picinpar}
\usepackage{amsmath}
\usepackage{stfloats}
\usepackage{url}
\usepackage{flushend}
\usepackage{colortbl}
\usepackage{xcolor,soul,framed} 
\colorlet{shadecolor}{yellow}
\usepackage{multirow}
\usepackage{pifont}
\usepackage{color,soul}
\usepackage{alltt}
\usepackage[hidelinks]{hyperref}
\usepackage{enumerate}
\usepackage{siunitx}
\usepackage{breakurl}
\usepackage{epstopdf}
\usepackage{pbox}
\usepackage{amssymb}
\usepackage{amsthm}
\usepackage{comment}
\usepackage{amsthm}

\usepackage{subfigure}
\usepackage{algorithm}
\usepackage{algpseudocode}

\usepackage{autobreak}
\usepackage[percent]{overpic} 
\usepackage{rotating}
\graphicspath{{../figures/}}
\DeclareGraphicsExtensions{.eps,.eps_tex,.pdf,.jpeg,.png}
\definecolor{limegreen}{rgb}{0.2, 0.8, 0.2}
\definecolor{forestgreen}{rgb}{0.13, 0.55, 0.13}
\definecolor{greenhtml}{rgb}{0.0, 0.5, 0.0}
\definecolor{skyblue}{rgb}{0.53, 0.81, 0.92}
\definecolor{lightgray}{rgb}{0.83, 0.83, 0.83}
\definecolor{gray}{rgb}{0.75, 0.75, 0.75}
\definecolor{darkgray}{rgb}{0.66, 0.66, 0.66}
\usepackage{mathtools}
\colorlet{shadecolor}{yellow!255}
\usepackage{lineno}
\usepackage{picins}  
\usepackage{url}  
\modulolinenumbers[5]

\makeatletter
\def\ps@pprintTitle{%
   \let\@oddhead\@empty
   \let\@evenhead\@empty
   \def\@oddfoot{\reset@font\hfil\thepage\hfil}
   \let\@evenfoot\@oddfoot
}
\makeatother

\begin{document}
\begin{frontmatter}

\title{Workspace Analysis and Optimal Design of Cable-Driven Parallel Robots via Auxiliary Counterbalances\tnoteref{t1}}
\tnotetext[t1]{This work was supported in part by the Natural Sciences and Engineering Research Council of Canada (NSERC).}

\author{Ronghuai Qi\corref{cor1}} \ead{r5qi@uwaterloo.ca}
\author{Hamed Jamshidifar\corref{}} \ead{hjamshid@uwaterloo.ca}
\author{Amir Khajepour\corref{}} \ead{a.khajepour@uwaterloo.ca}

\cortext[cor1]{Corresponding author}
\address{Department of Mechanical and Mechatronics Engineering, University of Waterloo, Waterloo, ON N2L 3G1, Canada}

\begin{abstract}
Cable-driven parallel robots (CDPRs) are widely investigated and applied in the worldwide; however, traditional configurations make them to be limited in reaching their maximum workspace duo to constraints such as the maximum allowable tensions of cables. In this paper, we introduce auxiliary counterbalances to tackle this problem and focus on workspace analysis and optimal design of CDPRs with such systems. Besides, kinematics, dynamics, and parameters optimization formulas and algorithm are provided to maximize the reachable workspace of CDPRs. Case studies for different configurations are presented and discussed. Numerical results suggest the effectiveness of the aforementioned approaches, and the obtained parameters can also be applied for actual CDPRs design.
\end{abstract}

\end{frontmatter}

\section{Introduction}\label{sec:J8_Introduction}
Cable-driven parallel robots (CDPRs) are an important type of industrial robot. Their configurations usually bear a resemblance to parallel manipulators. In these robots, rigid links are replaced with cables (e. g., a typical CDPR - NIST RoboCrane in{~}\cite{Albus1989,Albus1992}) in order to reduce their weight. It also eliminates the need for these revolute joints. These features allow the mobile platform to reach high motion accelerations in large workspaces. Due to these benefits, CDPRs are widely used in industry, rehabilitation, and other fields. For instance,  researchers{~}\cite{Mendez2014,Jamshidifar2020,Jamshidifar2018thesis} developed a CDPR, where the mobile platform is driven by two sets of upper cables and two sets of lower cables. They focused on the in-plane vibration control of the CDPR. However, owing to the effect of gravity on the cable-driven mobile platforms, conventional CDPRs cannot reach the top positions of the desired workspace since cable tensions may reach their maximum allowable values, and this problem widely exists in CDPRs.

To overcome the workspace limitation problem, researchers{~}\cite{Jamshidifar2020,Jamshidifar2018thesis,R.Qi2019j4,Qi2019thesis,T.Arai1999,H.Osumi2000,M.Bamdad2015,M.Gouttefarde2017,J.S.Albus1989,J.S.Albus2003,GmbH,SKYCAM_LLC} have attempted to mount manipulators or tools on the CDPRs. Literature shows that existing research and applications prefer to affix a robot arm upside down to the bottom of a CDPR’s platform{~}\cite{T.Arai1999,H.Osumi2000,M.Bamdad2015,M.Gouttefarde2017,J.S.Albus1989,J.S.Albus2003,GmbH,SKYCAM_LLC}. But this configuration cannot effectively improve the reachable workspace on the top space. The authors{~}\cite{R.Qi2019j4,Qi2019thesis} developed hybrid cable-driven robots (HCDRs), in which robot arms are mounted on the top of the CDPRs to deal with this problem. The experimental CDPR was based on the existing planar CDPR{~}\cite{Mendez2014}, in which a mobile platform is driven by two sets of upper cables and two sets of lower cables. Jamshidifar{~}\cite{Jamshidifar2018thesis} used this platform for studying the rigid body and in-plane vibration control of CDPR. Rushton{~}\cite{Rushton2016} introduced two pendulums to eliminate out-of-plane vibrations. The HCDRs can overcome the shortcomings of CDPRs and serial robots as well as aggregate their advantages. When a serial robot is mounted on a mobile platform, the two constitute a new coupled system, leading some new problems, e.g., only controlling the mobile platform or the serial robot may not guarantee the position accuracy of the end-effector. Besides, another major challenge in the utilization of these systems is maintaining the appropriate cable tensions and stiffness for the robot. This requires the development of kinematic and dynamic models, stiffness optimization, and controllers for HCDRs. Some research has been carried out to solve these problems: for kinematic and dynamic modeling, existing research mainly focuses on rigid serial robots{~}\cite{P.R.Pagilla2004} and rigid/flexible parallel robots{~}\cite{Mendez2014,Jamshidifar2018thesis,D.Lau2013,N.Mostashiri2018,C.Viegas2017,H.D.Taghirad2011,Z.Mu2020,M.J.Otis2009,M.Chen2018}. Some useful methods were studied to solve the redundancy and stiffness optimization problems, such as the minimum 2-norm of cable tensions{~}\cite{Mendez2014,Jamshidifar2018thesis} and stiffness maximization in the softest direction{~}\cite{Jamshidifar2018thesis}. However, the former one cannot make all cable tensions be positive; the latter one is complicated. Additionally, the moving robot arm also generates reaction forces acting on the mobile platform, resulting in mobile platform vibrations. In short, it is challenging to achieve the goal of minimizing the vibrations and increasing the position accuracy of the end-effector simultaneously. Hence, mounting manipulators and tools on the CDPRs makes the overall system complicated and results in vibrations and other problems above. These problems do not contribute to the improvement of the workspace.

Regarding the workspace analysis and optimal design, researchers developed numerical generation methods. For instance, Pham et al.{~}\cite{C.B.Pham2004} proposed a simplex search method to obtain the available workspace, but the computation efficiency is not high. Taking a planar CDPR as the object, Li{~}\cite{Y.Li2006} and Bolboli et al.{~}\cite{J.Bolboli2019} developed GA-based methods to studied the optimal design and workspace analysis, but whether the design methodology can be extended to other CDPRs are still ongoing. Tang{~}\cite{X.Tang2013} proposed a workspace quality index to analyze the workspace. Pusey{~}\cite{J.Pusey2004} et al. studied the design and workspace analysis of a 6-6 CDPR by using the set of points of the workspace volume. Merlet{~}\cite{J.P.Merlet2016} investigated the workspace of suspended CDPRs using straight-line cables, straight-line linear elastic cables, and sagging cables. But the author did not provide a clear analytical solution to determine the border equation for sagging cable since it was much more complicated in this situation. However, such techniques cannot tackle the problems for the configuration in this paper. Although the authors in{~}\cite{Q.Duan2015} studied the effects on the wrench-feasible workspace of CDPRs by adding springs, they provided an optimization equation to find feasible spring parameters, but not the optimal parameters of CDPRs. Besides, they used the existing null space of the structure matrix method in{~}\cite{S.R.Oh2005} to solve the workspace optimization problem, i.e., they didn't contribute to the improvement of the workspace optimization algorithm. The numerical results showed that adding springs on a planar CDPR didn’t increase (i.e., just change) the reachable workspace. In order to tackle the problems above, in this paper, we present workspace analysis and optimal design of CDPRs via auxiliary counterbalances. The main contributions are as follows:
\begin{enumerate}
    \item Auxiliary counterbalance systems are introduced to increase the reachable workspace of CDPRs.
    \item Modeling and parameters optimization approaches for CDPRs workspace extension are investigated with such systems.
    \item Different configurations are proposed and compared for optimal design.
    \item Case studies are conducted to verify the effectiveness of aforementioned approaches.
\end{enumerate}

The rest of this paper is organized as below: in{~}\autoref{sec:J8_Modeling}, we focus on system modeling, including kinematics and dynamics. In{~}\autoref{sec:J8_WorkspaceAnalysisMethod}, we develop a new workspace analysis method. Then, in{~}\autoref{sec:J8_NumericalResults}, we use specific CDPR numerical results to verify the proposed methods. Finally, in{~}\autoref{sec:J8_Conclusions}, we summarize the conclusions and further work.

\begin{figure}[!t]\centering
	\includegraphics[width=150mm]{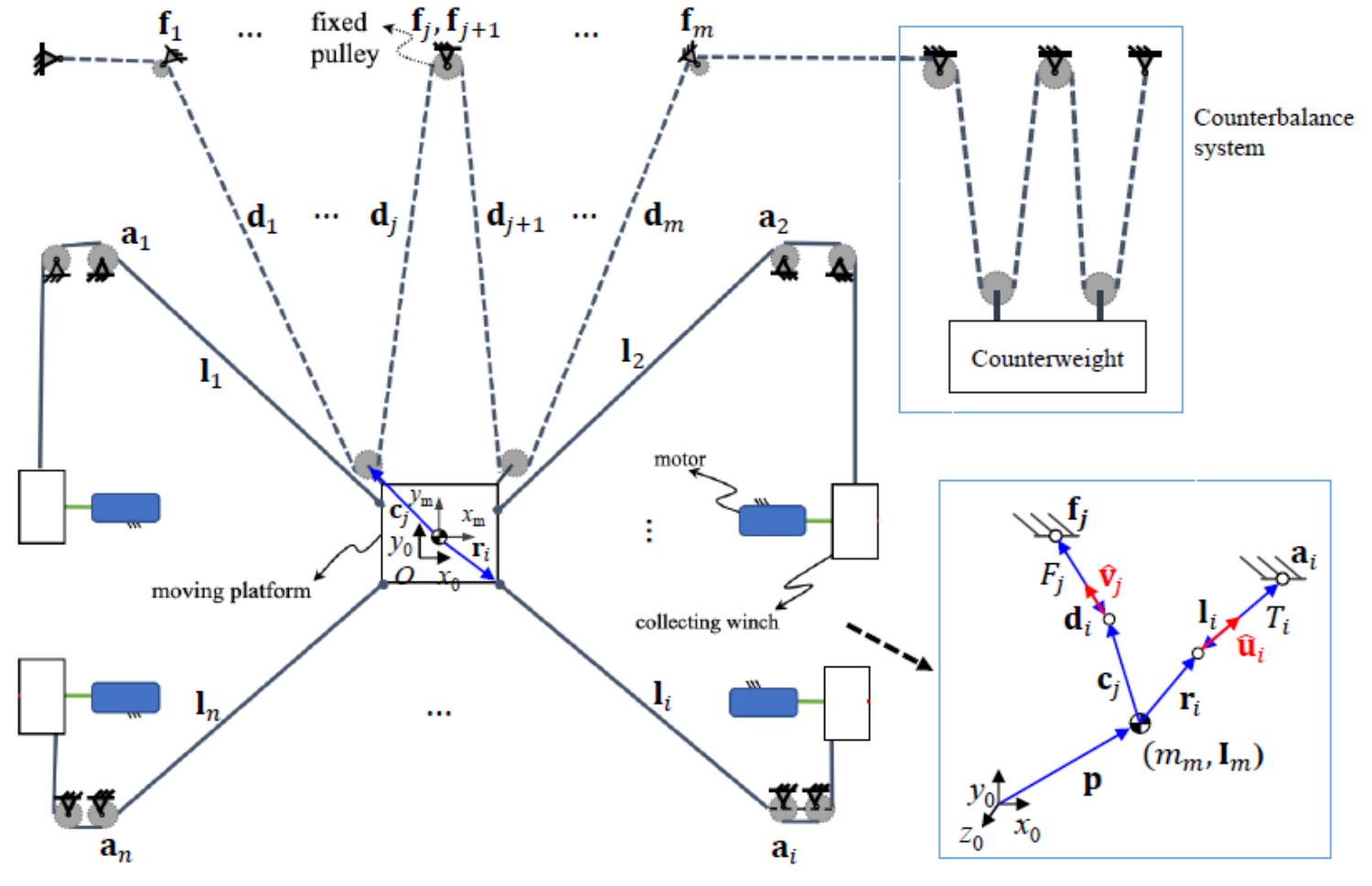}
	\caption{A general CDPR with a genetic auxiliary counterbalance system.}
	\label{fig:J8_Figure_1}
\end{figure}

\section{Modeling}\label{sec:J8_Modeling}
\subsection{General Configuration}\label{subsec:J8_GeneralConfiguration}
Consider a general cable-driven parallel robot, i.e., a moving platform that is actuated by multi-cables (see{~}\autoref{fig:J8_Figure_1}) and all cables are assumed to be straight and massless. To increase the reachable workspace, we introduce auxiliary counterbalances. The configuration and coordinate assignment of the CDPR with a genetic auxiliary counterbalance system are shown in{~}\autoref{fig:J8_Figure_1}, in which the CDPR has $n$ ($n \in{\mathbb{N}}$) driven cables (actuated by active winches), and $m$ ($m \in{\mathbb{N}}$) cables (actuated by a counterbalance system) connected to the auxiliary fixed pulleys. The genetic auxiliary counterbalance system consists of muti fixed pulleys and a counterweight, in which the counterweight can be replaced by a hydraulic system or other similar counterbalances. In this paper, we focus on using the counterweight for modeling and analysis. ${\bf{a}}_i$ ($\{\forall \; i \in{\mathbb{N}} :1 \le i \le n\}$) and ${\bf{f}}_j$ ($\{\forall \; j \in{\mathbb{N}} :1 \le j \le m\}$) are position vectors of the $i$-th cable and $j$-th cable, respectively, with respect to the global coordinate frame $\rm{X_0Y_0Z_0}$. ${\bf{r}}_i$ and ${\bf{c}}_j$ are body-fixed position vectors of the $i$-th anchor and $j$-th pulley on the moving platform, respectively. Other geometrical parameters such as ${\bf{l}}_i$ and ${\bf{d}}_j$ are also shown in{~}\autoref{fig:J8_Figure_1}. $m_m$ and ${\bf{I}}_m$ respectively denote the mass and moment of inertia of the mobile platform. Regarding the general configuration in{~}\autoref{fig:J8_Figure_1}, we develop the system modeling in the following sections.

\subsection{Kinematics}\label{subsec:J8_Kinematics}
Given the position vector of the mobile platform ${\bf{p}}$, ${\bf{r}}_i$, ${\bf{a}}_i$, ${\bf{c}}_j$, and ${\bf{f}}_j$ (see \autoref{fig:J8_Figure_1}. Note: if these vectors are given with respect to their local frames, one can use the rotation matrix in{~}\cite{R.Qi2019j4} to convert them from local to global coordinates), we can derive the kinematics as follows: the $i$-th cable length vector is calculated as
\begin{align}
    {{\bf{l}}_i} = {\bf{p}} + {{\bf{r}}_i} - {{\bf{a}}_i},\quad i = 1,2, \cdots ,n.
    \label{eq:J8_kine1}
\end{align}
Then, we get
\begin{align}
    {l_i} &= \left\| {{{\bf{l}}_i}} \right\| = \left\| {{\bf{p}} + {{\bf{r}}_i} - {{\bf{a}}_i}} \right\|,
    \label{eq:J8_kine2}\\
    {{\bf{\hat u}}_i} &=  - \frac{{{{\bf{l}}_i}}}{{{l_i}}},\label{eq:J8_kine3}
\end{align}
with ${l_i}$ and ${{\bf{\hat u}}_i}$ denoting the cable length and unit vector of the $i$-th cable, respectively. Similarly, for the auxiliary fixed pulleys, we have
\begin{align}
    {{\bf{d}}_j} &= {\bf{p}} + {{\bf{c}}_j} - {{\bf{f}}_j},\quad j = 1,2, \cdots ,m,\label{eq:J8_kine4}\\
    {d_j} &= \left\| {{{\bf{d}}_j}} \right\| = \left\| {{\bf{p}} + {{\bf{c}}_j} - {{\bf{f}}_j}} \right\|,\label{eq:J8_kine5}\\
    {{\bf{\hat v}}_j} &=  - \frac{{{{\bf{d}}_j}}}{{{d_j}}},\label{eq:J8_kine6}
\end{align}
where ${{\bf{d}}_j}$ and ${d_j}$ represent the $j$-th cable length vector and cable length, respectively. ${{\bf{\hat v}}_j}$ denotes the unit vector of the $j$-th cable connected to the corresponding auxiliary fixed pulley. Differentiating{~}\eqref{eq:J8_kine1} and{~}\eqref{eq:J8_kine4}, we get
\begin{align}
     - {{\dot l}_i}{{{\bf{\hat u}}}_i} - {{\bf{\omega }}_i} \times ({l_i}{{{\bf{\hat u}}}_i}) &= {\bf{\dot p}} + {{\bf{\omega }}_m} \times {{\bf{r}}_i}\\
    {{\dot l}_i} &=  - {{{\bf{\hat u}}}_i}({\bf{\dot p}} + {{\bf{\omega }}_m} \times {{\bf{r}}_i} + {{\bf{\omega }}_i} \times ({l_i}{{{\bf{\hat u}}}_i}))\\
     &=  - [\begin{array}{*{20}{c}}
    {{\bf{\hat u}}_i^T}&{{{({{\bf{r}}_i} \times {{{\bf{\hat u}}}_i})}^T}}
    \end{array}]\left[ {\begin{array}{*{20}{c}}
    {{\bf{\dot p}}}\\
    {{{\bf{\omega }}_m}}
    \end{array}} \right],
    \label{eq:J8_kine7}
\end{align}
and then, the vector of cable length velocity can be described as
\begin{align}
    \left[ {\begin{array}{*{20}{c}}
    {{{\dot l}_1}}\\
    {{{\dot l}_2}}\\
     \vdots \\
    {{{\dot l}_n}}
    \end{array}} \right] =  - \underbrace {\left[ {\begin{array}{*{20}{c}}
    {\begin{array}{*{20}{c}}
    {{\bf{\hat u}}_1^T}&{{{({{\bf{r}}_1} \times {{{\bf{\hat u}}}_1})}^T}}
    \end{array}}\\
    {\begin{array}{*{20}{c}}
    {{\bf{\hat u}}_2^T}&{{{({{\bf{r}}_2} \times {{{\bf{\hat u}}}_2})}^T}}
    \end{array}}\\
     \cdots \\
    {\begin{array}{*{20}{c}}
    {{\bf{\hat u}}_n^T}&{{{({{\bf{r}}_n} \times {{{\bf{\hat u}}}_n})}^T}}
    \end{array}}
    \end{array}} \right]}_{ \buildrel \Delta \over = {{\bf{J}}_l}}\left[ {\begin{array}{*{20}{c}}
    {{\bf{\dot p}}}\\
    {{{\bf{\omega }}_m}}
    \end{array}} \right],
    \label{eq:J8_kine8}
\end{align}
where ${{\bf{\omega }}_i}$ represents the angle velocity of the $i$-th cable, ${\bf{\dot p}}$ and ${{\bf{\omega }}_m}$ are linear and angle velocities of the mobile platform. ${{\bf{J}}_l}$ denotes the Jacobian matrix of the CDPR. Similarly, for the auxiliary fixed pulleys, we also have
\begin{align}
    {\dot d_j} &=  - [\begin{array}{*{20}{c}}
    {{\bf{\hat v}}_j^T}&{{{({{\bf{c}}_j} \times {{{\bf{\hat v}}}_j})}^T}}
    \end{array}]\left[ {\begin{array}{*{20}{c}}
    {{\bf{\dot p}}}\\
    {{{\bf{\omega }}_m}}
    \end{array}} \right],
    \label{eq:J8_kine9}\\
    \left[ {\begin{array}{*{20}{c}}
    {{{\dot d}_1}}\\
    {{{\dot d}_2}}\\
     \vdots \\
    {{{\dot d}_m}}
    \end{array}} \right] &=  - \underbrace {\left[ {\begin{array}{*{20}{c}}
    {\begin{array}{*{20}{c}}
    {{\bf{\hat v}}_1^T}&{{{({{\bf{c}}_1} \times {{{\bf{\hat v}}}_1})}^T}}
    \end{array}}\\
    {\begin{array}{*{20}{c}}
    {{\bf{\hat v}}_2^T}&{{{({{\bf{c}}_2} \times {{{\bf{\hat v}}}_2})}^T}}
    \end{array}}\\
     \cdots \\
    {\begin{array}{*{20}{c}}
    {{\bf{\hat v}}_m^T}&{{{({{\bf{c}}_m} \times {{{\bf{\hat v}}}_m})}^T}}
    \end{array}}
    \end{array}} \right]}_{ \buildrel \Delta \over = {{\bf{J}}_d}}\left[ {\begin{array}{*{20}{c}}
    {{\bf{\dot p}}}\\
    {{{\bf{\omega }}_m}}
    \end{array}} \right],
    \label{eq:J8_kine10}
\end{align}
where ${\dot d_j}$ means the $j$-th cable length velocity. ${{\bf{J}}_d}$ denotes the Jacobian matrix resulting from the auxiliary fixed pulleys. Finally, combining{~}\eqref{eq:J8_kine8} and{~}\eqref{eq:J8_kine10}, we can obtain
\begin{align}
    \left[ \begin{array}{l}
    \begin{array}{*{20}{c}}
    {{{\dot l}_1}}\\
    {{{\dot l}_2}}\\
     \vdots \\
    {{{\dot l}_n}}
    \end{array}\\
    \begin{array}{*{20}{c}}
    {{{\dot d}_1}}\\
    {{{\dot d}_2}}\\
     \vdots \\
    {{{\dot d}_m}}
    \end{array}
    \end{array} \right] =  - \underbrace {\left[ \begin{array}{l}
    \begin{array}{*{20}{c}}
    {\begin{array}{*{20}{c}}
    {{\bf{\hat u}}_1^T}&{{{({{\bf{r}}_1} \times {{{\bf{\hat u}}}_1})}^T}}
    \end{array}}\\
    {\begin{array}{*{20}{c}}
    {{\bf{\hat u}}_2^T}&{{{({{\bf{r}}_2} \times {{{\bf{\hat u}}}_2})}^T}}
    \end{array}}\\
     \cdots \\
    {\begin{array}{*{20}{c}}
    {{\bf{\hat u}}_n^T}&{{{({{\bf{r}}_n} \times {{{\bf{\hat u}}}_n})}^T}}
    \end{array}}
    \end{array}\\
    \begin{array}{*{20}{c}}
    {\begin{array}{*{20}{c}}
    {{\bf{\hat v}}_1^T}&{{{({{\bf{c}}_1} \times {{{\bf{\hat v}}}_1})}^T}}
    \end{array}}\\
    {\begin{array}{*{20}{c}}
    {{\bf{\hat v}}_2^T}&{{{({{\bf{c}}_2} \times {{{\bf{\hat v}}}_2})}^T}}
    \end{array}}\\
     \cdots \\
    {\begin{array}{*{20}{c}}
    {{\bf{\hat v}}_m^T}&{{{({{\bf{c}}_m} \times {{{\bf{\hat v}}}_m})}^T}}
    \end{array}}
    \end{array}
    \end{array} \right]}_{ \buildrel \Delta \over = {\bf{J}}}\left[ {\begin{array}{*{20}{c}}
    {{\bf{\dot p}}}\\
    {{{\bf{\omega }}_m}}
    \end{array}} \right],
    \label{eq:J8_kine8}
\end{align}
where ${\bf{J}}$ is the Jacobian matrix of the whole system.

\subsection{Dynamics}\label{subsec:J8_Dynamics}
Using the obtained results in{~}\autoref{subsec:J8_Kinematics} and Newton-Euler’s law, we can derive the equations of motion of the whole system as
\begin{align}
    \left[ {\begin{array}{*{20}{c}}
    {{m_m}{{{\bf{\dot v}}}_m}}\\
    {{{\bf{I}}_m}{{{\bf{\dot \omega }}}_m} + {{\bf{\omega }}_m} \times ({{\bf{I}}_m}{{\bf{\omega }}_m})}
    \end{array}} \right] + \left[ {\begin{array}{*{20}{c}}
    {{m_m}{{(0,g,0)}^T} + {{\bf{F}}_e}}\\
    {{{\bf{M}}_e}}
    \end{array}} \right]
     &= \left[ {\begin{array}{*{20}{c}}
    {\sum\limits_{i = 1}^n {({T_i}{{{\bf{\hat u}}}_i})}  + \sum\limits_{j = 1}^m {({F_j}{{{\bf{\hat v}}}_j})} }\\
    {\sum\limits_{i = 1}^n {({T_i}{{\bf{r}}_i} \times {{{\bf{\hat u}}}_i}) + \sum\limits_{j = 1}^m {({F_j}{{\bf{c}}_j} \times {{{\bf{\hat v}}}_j})} } }
    \end{array}} \right] \nonumber\\
     &= \left[ {\begin{array}{*{20}{c}}
    {{{{\bf{\hat u}}}_1}}&{{{{\bf{\hat u}}}_2}}& \cdots &{{{{\bf{\hat u}}}_n}}\\
    {{{\bf{r}}_1} \times {{{\bf{\hat u}}}_1}}&{{{\bf{r}}_2} \times {{{\bf{\hat u}}}_2}}& \cdots &{{{\bf{r}}_n} \times {{{\bf{\hat u}}}_n}}
    \end{array}} \right. \nonumber\\
    & \quad\; \left. {\begin{array}{*{20}{c}}
    {{{{\bf{\hat v}}}_1}}&{{{{\bf{\hat v}}}_2}}& \cdots &{{{{\bf{\hat v}}}_m}}\\
    {{{\bf{c}}_1} \times {{{\bf{\hat v}}}_1}}&{{{\bf{c}}_2} \times {{{\bf{\hat v}}}_2}}& \cdots &{{{\bf{c}}_m} \times {{{\bf{\hat v}}}_m}}
    \end{array}} \right]\left[ {\begin{array}{*{20}{c}}
    {\bf{T}}\\
    {\bf{F}}
    \end{array}} \right] \nonumber\\
     &= {{\bf{J}}^T}\left[ {\begin{array}{*{20}{c}}
    {\bf{T}}\\
    {\bf{F}}
    \end{array}} \right],
    \label{eq:J8_dyn1}
\end{align}
where ${\bf{T}} \buildrel \Delta \over = [{T_1},{T_2}, \cdots {T_n}]$ is the cable tension vector of the CDPR. ${\bf{F}} \buildrel \Delta \over = [{F_1},{F_2}, \cdots {F_m}]$ represent the cable tension vector from auxiliary fixed pulleys. ${\bf{F}_e}$ and ${\bf{M}_e}$ are external forces and moments, respectively. Then, Eq.{~}\eqref{eq:J8_dyn1} can be arranged as
\begin{align}
    \begin{array}{l}
    \underbrace {\left[ {\begin{array}{*{20}{c}}
    {{m_m}}&{}\\
    {}&{{{\bf{I}}_m}}
    \end{array}} \right]}_{ \buildrel \Delta \over = {\bf{M}}}\underbrace {\left[ {\begin{array}{*{20}{c}}
    {{{{\bf{\dot v}}}_m}}\\
    {{{{\bf{\dot \omega }}}_m}}
    \end{array}} \right]}_{ \buildrel \Delta \over = {\bf{\ddot q}}} + \underbrace {\left[ {\begin{array}{*{20}{c}}
    {\bf{0}}\\
    {{{\bf{\omega }}_m} \times ({{\bf{I}}_m}{{\bf{\omega }}_m})}
    \end{array}} \right]}_{ \buildrel \Delta \over = {\bf{C(q,\dot q)\dot q}}}
    + \underbrace {\left[ {\begin{array}{*{20}{c}}
    {{m_m}{{(0,g,0)}^T}}\\
    {\bf{0}}
    \end{array}} \right]}_{ \buildrel \Delta \over = {\bf{G(q)}}} + \left[ {\begin{array}{*{20}{c}}
    {{{\bf{F}}_e}}\\
    {{{\bf{M}}_e}}
    \end{array}} \right] = {{\bf{J}}^T}\left[ {\begin{array}{*{20}{c}}
    {\bf{T}}\\
    {\bf{F}}
    \end{array}} \right],
    \end{array}
    \label{eq:J8_dyn2}
\end{align}
where ${\bf{q}},{\bf{\dot q}},{\bf{\ddot q}}$ represent the vectors of generalized coordinates, velocities, and accelerations, respectively. $\bf{M}$ denotes the inertia matrix, $\bf{C(q, \dot q)}$ represents the combined Coriolis and centripetal matrix, and $\bf{G(q)}$ denotes the gravitational vector, respectively.

\section{Workspace Analysis Method}\label{sec:J8_WorkspaceAnalysisMethod}
To analyze the workspace, here, we ignore external force ${\bf{F}_e}$ and moment ${\bf{M}_e}$ for simplification. For an arbitrary equilibrium point, i.e., ${\bf{\ddot q}} = {\bf{\dot q}} = {\bf{0}}$, Eq.{~}\eqref{eq:J8_kine8} can be simplified as
\begin{align}
    {\bf{G(q)}} = {{\bf{J}}^T}\left[ {\begin{array}{*{20}{c}}
    {\bf{T}}\\
    {\bf{F}}
    \end{array}} \right] = \left[ {\begin{array}{*{20}{c}}
    {{\bf{J}}_l^T}&{{\bf{J}}_d^T}
    \end{array}} \right]\left[ {\begin{array}{*{20}{c}}
    {\bf{T}}\\
    {\bf{F}}
    \end{array}} \right] = {\bf{J}}_l^T{\bf{T}} + {\bf{J}}_d^T{\bf{F}}.
    \label{eq:J8_workspaceanalymethod1}
\end{align}
Defining ${\bf{u}} \buildrel \Delta \over = {\bf{G(q)}} - {\bf{J}}_d^T{\bf{F}}$ as an input vector, then we get
\begin{align}
    {\bf{u}} = {\bf{G(q)}} - {\bf{J}}_d^T{\bf{F}} = {\bf{J}}_l^T{\bf{T}}.
    \label{eq:J8_workspaceanalymethod2}
\end{align}
For a general case, Eq.{~}\eqref{eq:J8_workspaceanalymethod2} with constraints can be arranged as
\begin{align}
    {\bf{T}} = {({\bf{J}}_l^T)^ + }{\bf{u}} + {\bf{\alpha }}{\rm{Null}}({\bf{J}}_l^T),\quad 0 < {T_{i\min }} \le {T_i} \le {T_{i\max }},
    \label{eq:J8_workspaceanalymethod3}
\end{align}
where ${({\bf{J}}_l^T)^ + }$ and ${\rm{Null}}({\bf{J}}_l^T)$ represent pseudoinverse and null space of matrix ${\bf{J}}_l^T$, respectively. ${\bf{\alpha }}$ denotes a variable that can adjust $\bf{T}$ to satisfy with the constraints{~}\cite{S.Behzadipour2005}. Eq.{~}\eqref{eq:J8_workspaceanalymethod3} provides a genetic approach, but there may exist multi solutions and it is not easy to solve this problem for different robot configurations by using{~}\eqref{eq:J8_workspaceanalymethod3}. For the configuration in{~}\autoref{sec:J8_WorkspaceAnalysisMethod}, we propose another more effective method which can find a unique solution (let the rotation of the moving platform be zero). First, suppose
\begin{align}
    {{\bf{A}}_l} = [{{\bf{A}}_{l1}},{{\bf{A}}_{l2}},{{\bf{A}}_{l3}},{{\bf{A}}_{l4}}] \buildrel \Delta \over = {\bf{J}}_l^T,
    \label{eq:J8_workspaceanalymethod4}
\end{align}
where ${\bf{A}}_{l1}$ is the sub-vector of ${{\bf{A}}_l}$. In addition, the redundancy resolution problem resulting from multi-cables can be solved as follows: since the equivalent four-cable planar CDPR has one DOR, then is redefined as a new $3 \times 4$ matrix. One can restrict one of cable tensions $T_i\;(i=1,2,3,4)$ to the maximum allowable value ${T_{i\max }}$. In this case, the redundancy resolution and cable tensions are described as
\begin{align}
    \left\{ \begin{array}{r}
    {[{T_2},{T_3},{T_4}]^T} = {[{{\bf{A}}_{l2}},{{\bf{A}}_{l3}},{{\bf{A}}_{l4}}]^{ - 1}}({\bf{u}} - {{\bf{A}}_{l1}}{T_{1\max }}),\quad {\rm{for}}\;{T_1} = {T_{1\max }}\\
    {[{T_1},{T_3},{T_4}]^T} = {[{{\bf{A}}_{l1}},{{\bf{A}}_{l3}},{{\bf{A}}_{l4}}]^{ - 1}}({\bf{u}} - {{\bf{A}}_{l2}}{T_{2\max }}),\quad {\rm{for}}\;{T_2} = {T_{2\max }}\\
    {[{T_1},{T_2},{T_4}]^T} = {[{{\bf{A}}_{l1}},{{\bf{A}}_{l2}},{{\bf{A}}_{l4}}]^{ - 1}}({\bf{u}} - {{\bf{A}}_{l3}}{T_{3\max }}),\quad {\rm{for}}\;{T_3} = {T_{3\max }}\\
    {[{T_1},{T_2},{T_3}]^T} = {[{{\bf{A}}_{l1}},{{\bf{A}}_{l2}},{{\bf{A}}_{l3}}]^{ - 1}}({\bf{u}} - {{\bf{A}}_{l4}}{T_{4\max }}),\quad {\rm{for}}\;{T_4} = {T_{4\max }}
    \end{array} \right..
    \label{eq:J8_workspaceanalymethod5}
\end{align}

Eq.{~}\eqref{eq:J8_workspaceanalymethod5} includes four solutions without constraints. Combining these solutions and constraints of cables, we define the cost function (cables are assumed to be no elastic) as
\begin{align}
    \begin{array}{l}
    \begin{array}{*{20}{c}}
    {\Gamma  = \max }&{\left\{ {\left\| {{{\bf{T}}_{{\rm{opt1}}}}} \right\|,\left\| {{{\bf{T}}_{{\rm{opt2}}}}} \right\|,\left\| {{{\bf{T}}_{{\rm{opt3}}}}} \right\|,\left\| {{{\bf{T}}_{{\rm{opt4}}}}} \right\|} \right\}}
    \end{array}\\
    \begin{array}{*{20}{c}}
    {\quad \;\;\;\;\,{\rm{s}}{\rm{.t}}{\rm{.}}}&{{{\bf{T}}_{{\rm{opt1}}}} = {{[{T_{1\max }},{T_2},{T_3},{T_4}]}^T}}
    \end{array}\\
    \begin{array}{*{20}{c}}
    {\quad \;\;\;\quad \;\,\,}&{{{\bf{T}}_{{\rm{opt2}}}} = {{[{T_1},{T_{2\max }},{T_3},{T_4}]}^T}}
    \end{array}\\
    \quad \;\;\;\begin{array}{*{20}{c}}
    {\quad \;\,\,}&{{{\bf{T}}_{{\rm{opt3}}}} = {{[{T_1},{T_2},{T_{3\max }},{T_4}]}^T}}
    \end{array}\\
    \quad \;\;\;\begin{array}{*{20}{c}}
    {\quad \;\,\,}&{{{\bf{T}}_{{\rm{opt4}}}} = {{[{T_1},{T_2},{T_3},{T_{4\max }}]}^T}}
    \end{array}\\
    \quad \;\;\;\begin{array}{*{20}{c}}
    {\quad \;\,\,}&{0 \le {T_{k\min }} \le {T_k} \le {T_{k\max }},\quad k = 1,2,3,4,5},
    \end{array}
    \end{array}
    \label{eq:J8_workspaceanalymethod6}
\end{align}
where $\Gamma$ denotes the maximum stiffness of the CDPR. Clearly, after obtaining $\Gamma$, we can also get the corresponding optimal cable tension vector ${\bf{T}}_{{\rm{opt}}}^*$ (${{\bf{T}}_{{\rm{opt}}i}},i = 1,2,3,4$). From{~}\eqref{eq:J8_workspaceanalymethod2},{~}\eqref{eq:J8_workspaceanalymethod4},{~}\eqref{eq:J8_workspaceanalymethod5}, and{~}\eqref{eq:J8_workspaceanalymethod6}, we know that ${\bf{T}}_{{\rm{opt}}}^*$ is a function in terms of ${\bf{p}}(x, y)$, $w_p$ and $\bf{F}$. Since $\bf{F}$ is a vector including two elements (they are equal to each other), here, we use $T_5$ to represent one element. Then, for a desired position $(x, y)$, we can verify the reachability of this point using ${\bf{T}}_{{\rm{opt}}}^*({w_p},{T_5})$. $w_p$ and $T_5$ are position and force parameters to optimize the reachable workspace. We can also obtain more conclusions: given $x$, $y$, and $w_p$, ${\bf{T}}_{{\rm{opt}}}^*({w_p},{T_5})$ and $T_5$ are linear relative. In comparison with the method shown in{~}\cite{Rushton2016}, Eqs.{~}\eqref{eq:J8_workspaceanalymethod5} and{~}\eqref{eq:J8_workspaceanalymethod6} can ensure that all the cable tensions are positive. If cables are elastic, we can introduce cable length constraints into{~}\eqref{eq:J8_workspaceanalymethod6}:
\begin{align}
    \begin{array}{l}
    \begin{array}{*{20}{c}}
    {\Gamma  = \max }&{\left\{ {\left\| {{{\bf{T}}_{{\rm{opt1}}}}} \right\|,\left\| {{{\bf{T}}_{{\rm{opt2}}}}} \right\|,\left\| {{{\bf{T}}_{{\rm{opt3}}}}} \right\|,\left\| {{{\bf{T}}_{{\rm{opt4}}}}} \right\|} \right\}}
    \end{array}\\
    \begin{array}{*{20}{c}}
    {\quad \;\;\;\;\,{\rm{s}}{\rm{.t}}{\rm{.}}}&{{{\bf{T}}_{{\rm{opt1}}}} = {{[{T_{1\max }},{T_2},{T_3},{T_4}]}^T}}
    \end{array}\\
    \begin{array}{*{20}{c}}
    {\quad \;\;\;\quad \,\;\,}&{{{\bf{T}}_{{\rm{opt2}}}} = {{[{T_1},{T_{2\max }},{T_3},{T_4}]}^T}}
    \end{array}\\
    \quad \;\;\;\begin{array}{*{20}{c}}
    {\quad \;\,\,}&{{{\bf{T}}_{{\rm{opt3}}}} = {{[{T_1},{T_2},{T_{3\max }},{T_4}]}^T}}
    \end{array}\\
    \quad \;\;\;\begin{array}{*{20}{c}}
    {\quad \;\,\,}&{{{\bf{T}}_{{\rm{opt4}}}} = {{[{T_1},{T_2},{T_3},{T_{4\max }}]}^T}}
    \end{array}\\
    \quad \;\;\;\begin{array}{*{20}{c}}
    {\quad \;\,\,}&{0 \le {T_{k\min }} \le {T_k} \le {T_{k\max }},\quad k = 1,2,3,4,5}
    \end{array}\\
    \quad \;\;\;\begin{array}{*{20}{c}}
    {\quad \;\,\,}&{0 \le {l_{0k\min }} \le \frac{{{l_k}E{A_k}}}{{{T_k} + E{A_k}}} \le {l_{0k\max }},\quad k = 1,2,3,4,5},
    \end{array}
    \end{array}
    \label{eq:J8_workspaceanalymethod7}
\end{align}
where $E{A_k}$ represents the product of the modulus of elasticity and cross-sectional area of the $k$-th cable, and $l_k$ denotes $k$-th cable length. ${l_{0k\min }}$ and ${l_{0k\max }}$ are the minimum and maximum length of the $k$-th unstretched cable.

Additionally, we propose{~}\autoref{algorithm:J8_workspaceanalyalgorithm} to find the reachable workspace. Given the desired $x$, $y$, and $T_5$ and follow the steps in{~}\autoref{algorithm:J8_workspaceanalyalgorithm}, we can obtain a set of feasible position $(x,y)$ named reachable workspace. $x_{\min}$, $x_{\max}$, $y_{\min}$, $y_{\max}$ are kinematic constraints. In this paper, they are provided in \autoref{table:J8_CDPRParameters} for case studies.

\begin{algorithm*}[htb]
\caption {Calculation of reachable workspace.} \label{algorithm:J8_workspaceanalyalgorithm}
\begin{algorithmic}[1]
\Require {Given the desired $x$ and $y$ and $T_5$.}
\Ensure {Reachable workspace $x$, $y$.} \Comment{Position ${\bf{p}}(x,y)$.}
\State Given desired workspace $x_{\min}$, $x_{\max}$, $y_{\min}$, $y_{\max}$, and a set of $T_5$; \Comment{$x_{\min}$, $x_{\max}$, $y_{\min}$, $y_{\max}$ are kinematic constraints.}
\For{$x_{\min}$ to $x_{\max}$}
    \For{$y_{\min}$ to $y_{\max}$}
    \State {Using{~}\eqref{eq:J8_kine8} and{~}\eqref{eq:J8_kine10} to find ${\bf{J}}_l$ and ${\bf{J}}_d$, respectively;} 
    \State {Using{~}\eqref{eq:J8_workspaceanalymethod2} to find $\bf{u}$;}
    \State Substituting $\bf{u}$ and ${\bf{J}}_l$ into{~}\eqref{eq:J8_workspaceanalymethod4} and{~}\eqref{eq:J8_workspaceanalymethod5};
    \State Recording the current $x$, $y$ if{~}\eqref{eq:J8_workspaceanalymethod6} or{~}\eqref{eq:J8_workspaceanalymethod7} holds;
    \EndFor
\EndFor
\State {\bf Return} {Reachable workspace $x$, $y$.}
\end{algorithmic}
\end{algorithm*}

\section{Numerical Results and Discussion}\label{sec:J8_NumericalResults}
In this section, we conduct different scenarios for workspace analysis via the proposed models (in \autoref{sec:J8_Modeling}) and algorithms (in{~}\autoref{sec:J8_WorkspaceAnalysisMethod}). First, we provide a planar CDPR, which is actuated by four cables (see{~}\autoref{fig:J8_Figure_2}). All cables are assumed to be straight, massless, and non-elastic. In order to achieve the goal of workspace expansion and parameter optimization, we introduce one or more fixed pulleys on the mobile platform. Here, we propose four possible configurations for comparison (see{~}\autoref{fig:J8_Figure_2}). In{~}\autoref{fig:J8_Figure_2}, the arrangement of the cables are different, in which mounting a single pulley on the top side of the platform, mounting two pulleys on the top side of the platform, mounting two pulleys on the bottom side of the platform, and mounting two pulleys on the top side of the platform and one fixed pulley on the static frame are shown in{~}\autoref{fig:J8_Figure_2} (a)-(d), respectively. We can also add more numbers of auxiliary fixed pulleys (i.e., greater than two), but we found that other configurations do not help increase the reachable workspace. In this paper, we focus on one of them, i.e.,{~}\autoref{fig:J8_Figure_2} (a), for detailed analysis and case studies. Besides, in{~}\autoref{fig:J8_Figure_2} (a), the coordinate assignment of the proposed CDPR are given, with geometrical dimensions $w,h,{w_b},{w_p},{h_p},{h_{bp}},{w_{bp}},{h_{bu}},$ and ${h_{bl}}$. The global coordinate frame $\rm{X_0OY_0}$ is located at the center of the static frame. Two points ${\bf{c}}_1$ and ${\bf{c}}_2$ connected to auxiliary fixed pulleys ${\bf{f}}_1$ and ${\bf{f}}_2$ are located at the top center of the platform. The following vectors are defined as: ${\bf{r}}_1=[-w_b/2,h_{bl},0]^T$, ${\bf{r}}_2=[w_b/2,h_{bu},0]^T$, ${\bf{r}}_3=[w_b/2,-h_{bl},0]^T$, ${\bf{r}}_4=[-w_b/2,-h_{bu},0]^T$, ${\bf{c}}_1=[-w_{bp},h_{bp},0]^T$, ${\bf{c}}_2=[-w_{bp},h_{bp},0]^T$, ${\bf{a}}_1=[-w/2,h/2,0]^T$, ${\bf{a}}_2=[w/2,h/2,0]^T$, ${\bf{a}}_3=[w/2,-h/2,0]^T$, ${\bf{a}}_4=[-w/2,-h/2,0]^T$, ${\bf{f}}_1=[-w_p,h_p,0]^T$, ${\bf{f}}_2=[0,h_p,0]^T$, ${\bf{f}}_3=[0,h_p,0]^T$, and ${\bf{f}}_4=[w_p,h_p,0]^T$. Using the specific parameters $w,h,{w_b},{w_p},{h_p},{h_{bp}},{w_{bp}},{h_{bu}},$ and ${h_{bl}}$ provided in{~}\autoref{table:J8_CDPRParameters}, we can obtain the vectors above. Substituting these vectors in to the equations in{~}\autoref{sec:J8_Modeling} and{~}\autoref{sec:J8_WorkspaceAnalysisMethod}, we can find the corresponding items (in{~}\eqref{eq:J8_workspaceanalymethod1},{~}\eqref{eq:J8_workspaceanalymethod2}, and{~}\eqref{eq:J8_workspaceanalymethod4}) as below:
\begin{figure}[!t]\centering
	\includegraphics[width=163mm]{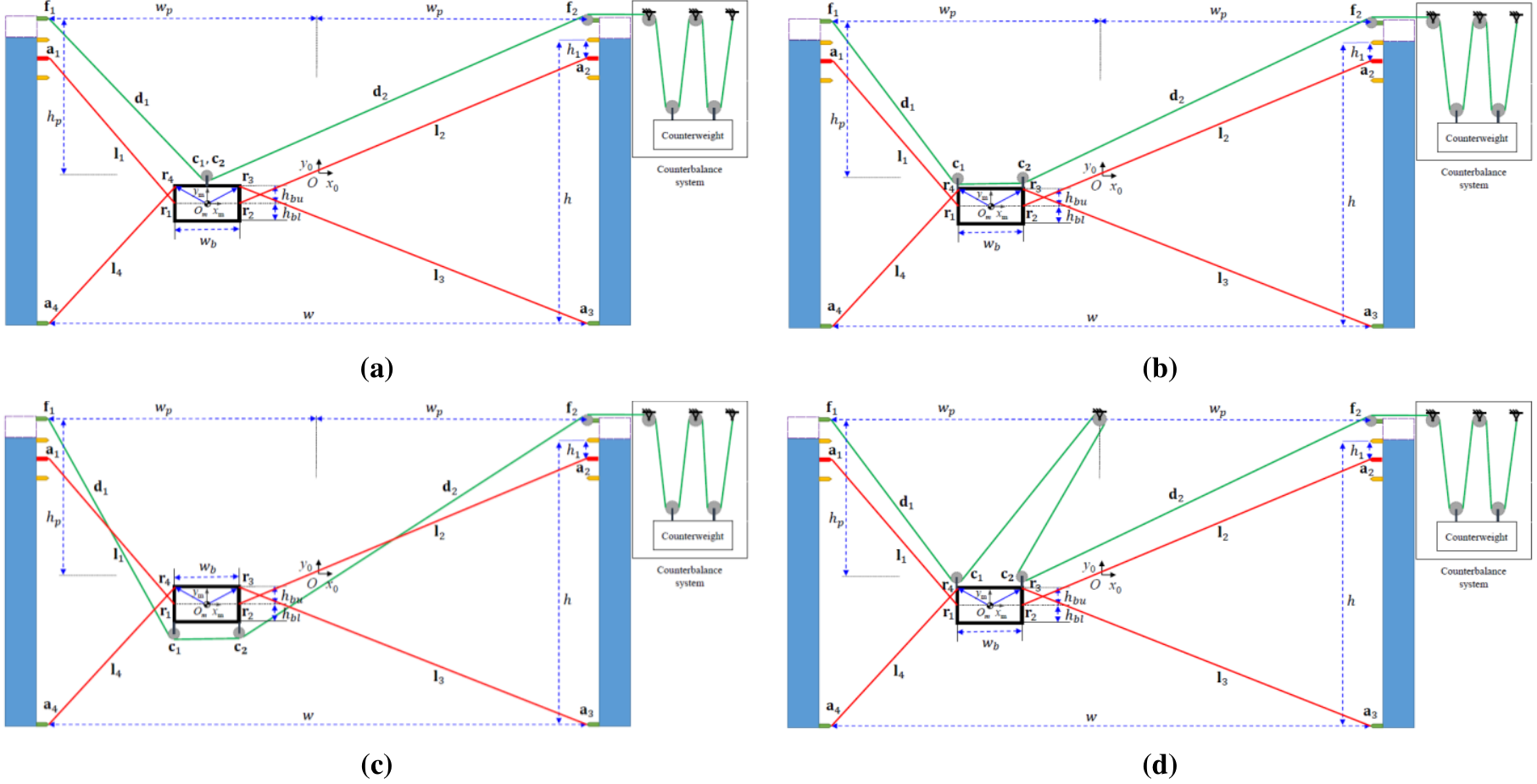}
	\caption{Possible robot configurations and coordinates assignment. (a) Mounting a single pulley on the top side of the platform. (b) Mounting two pulleys on the top side of the platform. (c) Mounting two pulleys on the bottom side of the platform. (d) Mounting two pulleys on the top side of the platform and one fixed pulley on the static frame.}
	\label{fig:J8_Figure_2}
\end{figure}

\begin{align}
    {\bf{G(q)}} = {[0,{m_m}g,0]^T},
    \label{eq:J8_numericalresult1}
\end{align}
where $m_m$ denotes the mass of the moving platform.
\begin{align}
    {\bf{F}} = [{T_5},{T_5}],
    \label{eq:J8_numericalresult2}
\end{align}
where $T_5$ represents the cable tension connected to a fixed pulley for optimal design. The structure matrix and Jacobian matrix are described as
\begin{align}
    {{\bf{A}}_l} = \left( {\begin{array}{*{20}{c}}
    { - \frac{{w - {w_b} + 2{\mkern 1mu} x}}{{2\chi }}}&{ - \frac{{{w_b} - w + 2{\mkern 1mu} x}}{{2{\mkern 1mu} \chi }}}&{ - \frac{{{w_b} - w + 2{\mkern 1mu} x}}{{2{\mkern 1mu} \delta }}}&{ - \frac{{w - {w_b} + 2{\mkern 1mu} x}}{{2\delta }}}\\
    { - \frac{{{h_1} - \frac{h}{2} + y}}{\chi }}&{ - \frac{{{h_1} - \frac{h}{2} + y}}{\chi }}&{ - \frac{{\frac{h}{2} + {h_{bu}} + y}}{\delta }}&{ - \frac{{\frac{h}{2} + {h_{bu}} + y}}{\delta }}\\
    {\frac{{{w_b}{\mkern 1mu} \left( {{h_1} - \frac{h}{2} + y} \right)}}{{2\chi }}}&{ - \frac{{{w_b}{\mkern 1mu} \left( {{h_1} - \frac{h}{2} + y} \right)}}{{2\chi }}}&{ - \frac{{2{\mkern 1mu} {h_{bu}}{\mkern 1mu} w + h{\mkern 1mu} {w_b} - 4{\mkern 1mu} {h_{bu}}{\mkern 1mu} x + 2{\mkern 1mu} {w_b}{\mkern 1mu} y}}{{4\delta }}}&{\frac{{2{\mkern 1mu} {h_{bu}}{\mkern 1mu} w + h{\mkern 1mu} {w_b} + 4{\mkern 1mu} {h_{bu}}{\mkern 1mu} x + 2{\mkern 1mu} {w_b}{\mkern 1mu} y}}{{4\delta }}}
    \end{array}} \right),
    \label{eq:J8_numericalresult3}
\end{align}
and
\begin{align}
    {{\bf{J}}_d} = {\left( {\begin{array}{*{20}{c}}
    { - \frac{{{w_p} + x}}{\eta }}&{\frac{{{w_p} - x}}{\kappa }}\\
    { - \frac{{{h_{bp}} - {h_p} + y}}{\eta }}&{ - \frac{{{h_{bp}} - {h_p} + y}}{\kappa }}\\
    {\frac{{{h_{bp}}{\mkern 1mu} \left( {{w_p} + x} \right)}}{\eta }}&{ - \frac{{{h_{bp}}{\mkern 1mu} \left( {{w_p} - x} \right)}}{\kappa }}
    \end{array}} \right)^T},
    \label{eq:J8_numericalresult4}
\end{align}
with $\chi  = \sqrt {{{({h_1} - \frac{h}{2} + y)}^2} + {{(\frac{w}{2} - \frac{{{w_b}}}{2} + x)}^2}}$, $\delta  = \sqrt {{{(\frac{h}{2} + {h_{bu}} + y)}^2} + {{(\frac{{{w_b}}}{2} - \frac{w}{2} + x)}^2}}$, $\eta  = \sqrt {{{({w_p} + x)}^2} + {{({h_{bp}} - {h_p} + y)}^2}}$, and $\kappa  = \sqrt {{{({w_p} - x)}^2} + {{({h_{bp}} - {h_p} + y)}^2}}$. $w_b$ is the other parameter (see{~}\autoref{fig:J8_Figure_2}) for optimization. Substituting{~}\eqref{eq:J8_numericalresult1}-\eqref{eq:J8_numericalresult4} back into{~}\eqref{eq:J8_workspaceanalymethod5}-\eqref{eq:J8_workspaceanalymethod6} and using{~}\autoref{algorithm:J8_workspaceanalyalgorithm}, we conduct the scenarios of workspace analysis and optimal design as follows:

\begin{table}[!t]
\renewcommand{\arraystretch}{1.3}
\caption{Parameters for case studies.}
\centering
\label{table:J8_CDPRParameters}
\centering
	\begin{tabular}{c c c c}
	\hline\hline \\[-3mm]
    Symbol & Value & Symbol & Value  \\[1.6ex] \hline
    $w$ & $28.0$ \si{\metre} & $h_p$ & $3.246$ \si{\metre} \\
    $h$ & $5.70$ \si{\metre} & $h_{bp}$ & $0.45$ \si{\metre} \\
    $w_b$ & $1.90$ \si{\metre} & $w_{bp}$ & $0.95$ \si{\metre} \\
    $h_1$ & $0.45$ \si{\metre} & $h_{bu}$ & $0.45$ \si{\metre} \\
    $[x_{\min}, x_{\max}]$ & $[-12.5, 12.5]$ \si{\metre} & $[y_{\min}, y_{\max}]$ &  $[-2.85, 2.15]$ \si{\metre}\\
    $m_m$ & \SI{300}{\kilogram} & $g$ & $9.81$ \si[per-mode=symbol]{\metre\per\second\squared}\\
    $T_{1\max},T_{2\max},T_{5\max}$ & \SI{16000}{\newton} & $T_{3\max},T_{4\max}$ & \SI{12000}{\newton}\\
    $T_{k\min },k = 1,2,3,4,5$ & \SI{0}{\newton} & \\
	\hline\hline
	\end{tabular}
\end{table}

\subsection{Scenario 1: Choosing $w_p$} \label{subsec:J8_ScenarioChooseWp}
\begin{figure}[!t]\centering
	\includegraphics[width=120mm]{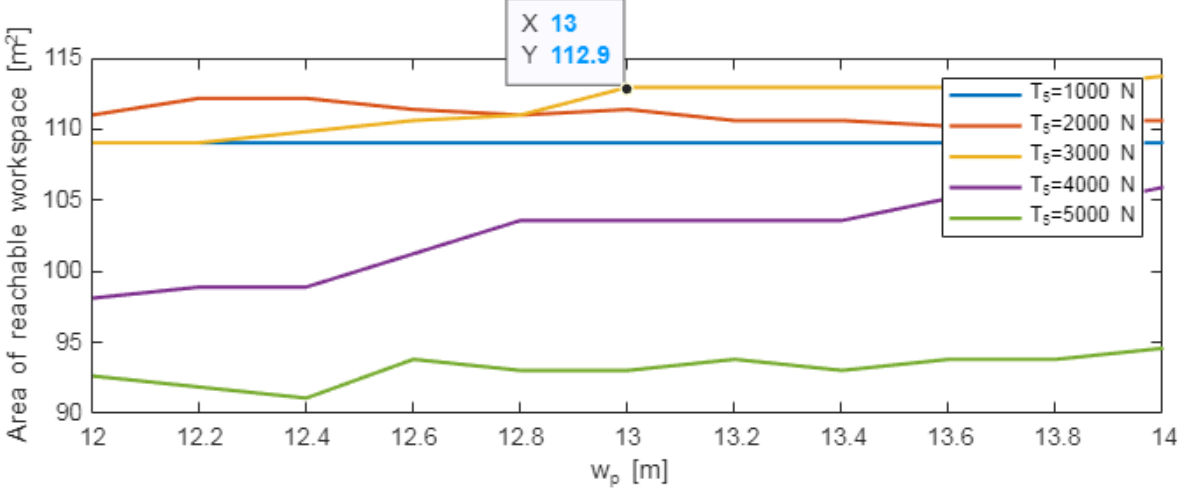}
	\caption{Area of reachable workspace versus $w_p$.}
	\label{fig:J8_Figure_3}
\end{figure}

In this scenario, we change the parameter $w_p$ to estimate the impact to the reachable workspace. Here, we conduct different conditions and let $T_5$ be 1000 N, 2000 N, 3000 N, 4000 N, and 5000 N for analysis. In each condition, $T_5$ is constant and the objective is find the optimal $w_p$. By carrying out{~}\autoref{algorithm:J8_workspaceanalyalgorithm}, we can obtain the results shown in{~}\autoref{fig:J8_Figure_3}. It is clear that $w_p$ equaling 13 m is a proper value that makes the area of reachable workspace be the largest.

\subsection{Scenario 2: Choosing $T_5$} \label{subsec:J8_ScenarioChooseT5}
\begin{figure}[!t]\centering
	\includegraphics[width=120mm]{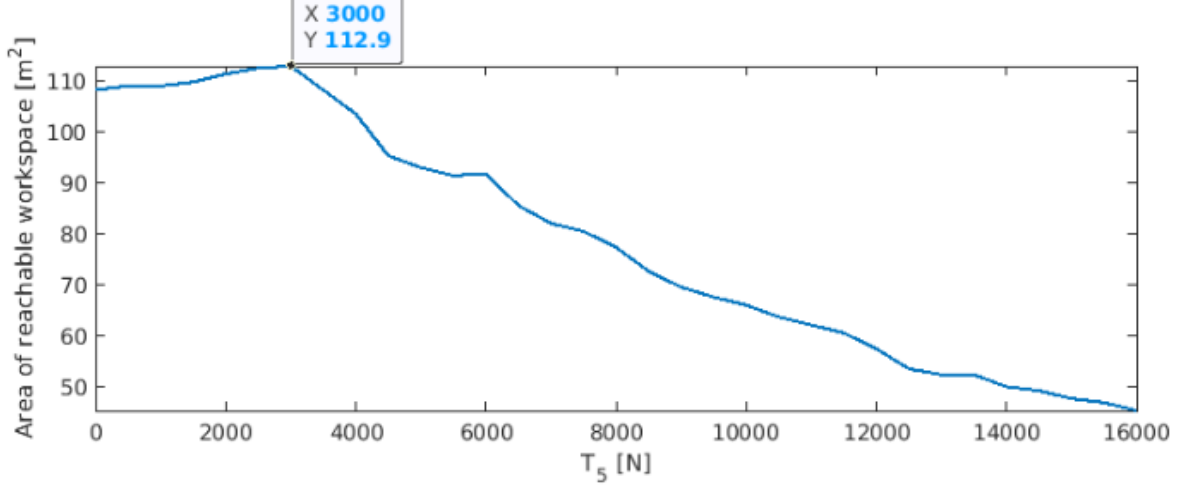}
	\caption{Area of reachable workspace versus $T_5$.}
	\label{fig:J8_Figure_4}
\end{figure}

In~\autoref{subsec:J8_ScenarioChooseWp}, we obtained the proper value of $w_p$ (13 m). Applying this value, we conduct a new scenario: changing $T_5$ to estimate its effect on the reachable workspace, and the result is shown in{~}\autoref{fig:J8_Figure_4}. The result reveals that when $T_5$ is equal to 3000 N, the workspace reaches the peak point, i.e., the optimal value. Increasing $T_5$ does not improve the reachable workspace. After obtaining $T_5$, the counterweight $M$ can be computed as $M={\varsigma}T_5$, where ${\varsigma}$ represents the number of cables in the counterbalance system.

\subsection{Discussion: Desired Workspace versus Reachable Workspace} \label{subsec:J8_DiscussionVSWorkspace}
\begin{figure}[!t]\centering
	\includegraphics[width=130mm]{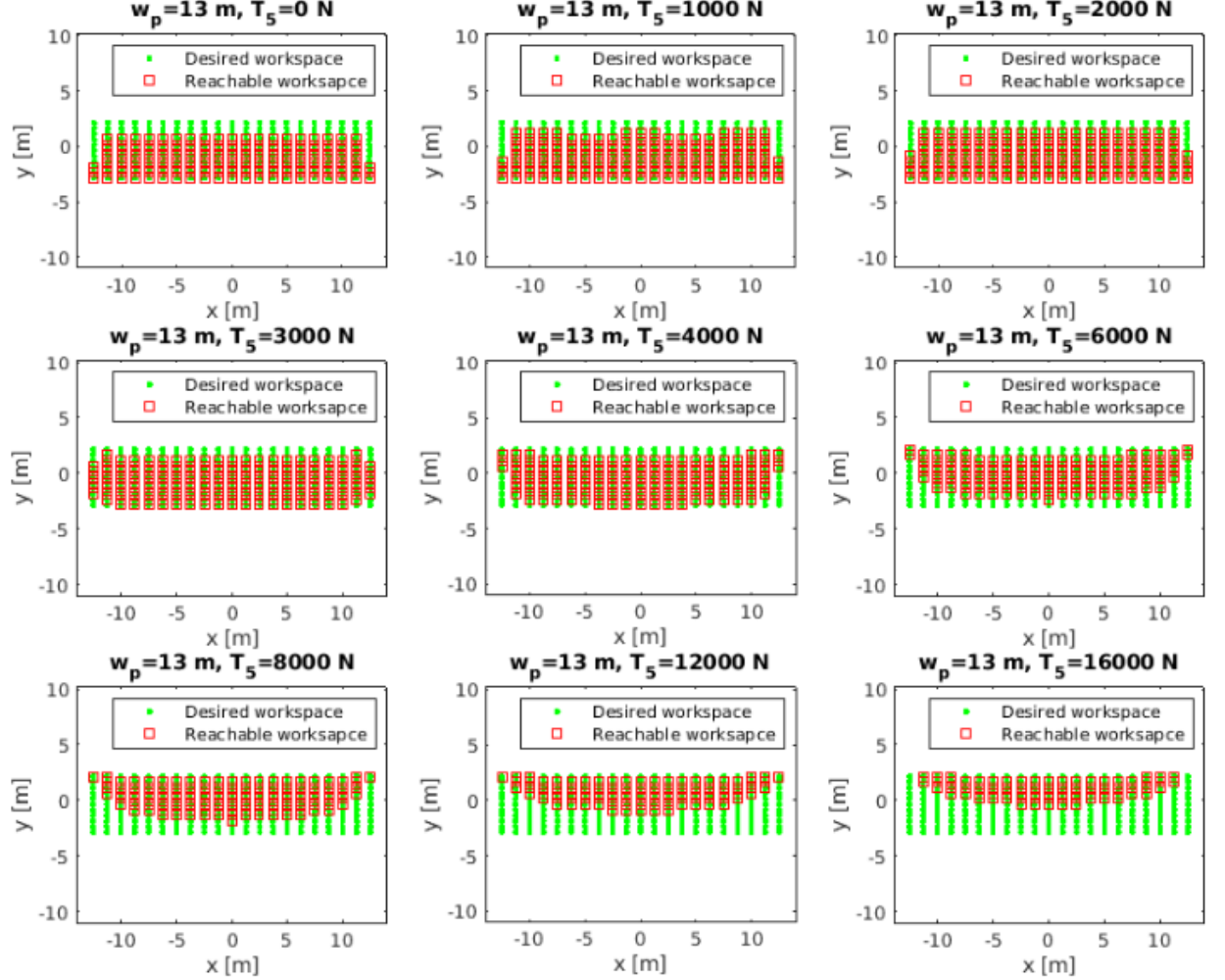}
	\caption{Desired workspace versus reachable workspace.}
	\label{fig:J8_Figure_5}
\end{figure}

In~\autoref{subsec:J8_ScenarioChooseWp} and~\autoref{subsec:J8_ScenarioChooseT5}, we obtained the optimal $w_p$ (13 m) and $T_5$ (3000 N) by maximizing the area of the reachable workspace. In actual system design, we are also interested in the surface of the workspace, e.g., the corner of the workspace. Here, we use an example of reachable workspace compared with the desired workspace (see{~}\autoref{fig:J8_Figure_5}) to illustrate this point. In{~}\autoref{fig:J8_Figure_5}, the results show that larger $T_5$ can increase reachable workspace on the top, but lose the reachable workspace on the bottom, especially on the bottom left and right. When $T_5$ is equal to 3000 N, the area of reachable workspace is the largest (112.9 ${\rm m^2}$);  when $T_5$ equals 2000 N, the area of the reachable workspace (112.9 ${\rm m^2}$) is smaller but the corners on the bottom can be covered. Hence, in real system design, we can choose the value of $T_5$ between 2000 N and 3000 N regarding the specific need. Ideally, we can use a hydraulic system to adjust $T_5$ to satisfy with maximizing the area of the workspace as well as covering the corners of the desired workspace.

\subsection{Discussion: Comparison of Different Robot Configurations} \label{subsec:J8_DiscussionConfig}
\begin{figure}[!t]\centering
	\includegraphics[width=163mm]{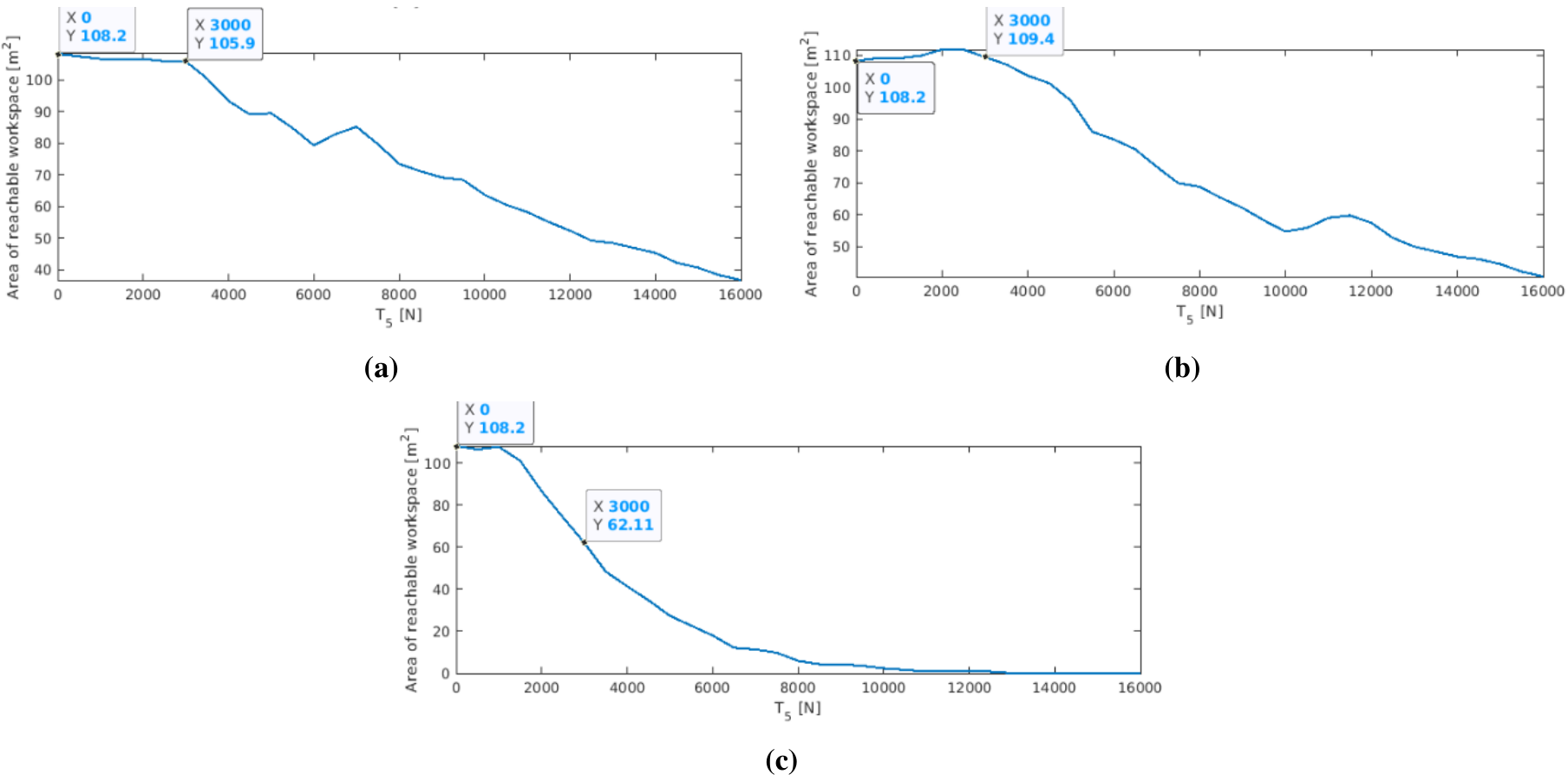}
	\caption{Area of reachable workspace of different robot configurations versus $T_5$. (a) Responses of the configuration{~}\autoref{fig:J8_Figure_2} (b). (b) Responses of the configuration{~}\autoref{fig:J8_Figure_2} (c). (c) Responses of the configuration{~}\autoref{fig:J8_Figure_2} (d).}
	\label{fig:J8_Figure_6}
\end{figure}

\autoref{fig:J8_Figure_2} provided four possible configurations, and then we focused on the one shown in{~}\autoref{fig:J8_Figure_2}(a) for detailed workspace expansion and parameter optimization. To compare the corresponding results with those of other configurations, we also conduct the following cases: To begin with, the area of reachable workspace of another three robot configurations are provided in{~}\autoref{fig:J8_Figure_6}, in which{~}\autoref{fig:J8_Figure_6} (a), (b), and (c) correspond to the configurations shown in{~}\autoref{fig:J8_Figure_2} (b), (c), and (d), respectively. Here, we use $w_p$ that is equal to 13 $\rm{m}$ and the scope of $T_5$ in all cases for easy comparison. Clearly, in comparison with the result shown in{~}\autoref{fig:J8_Figure_4}, when $T_5$ is equal to 3000 N, the areas of reachable workspace in{~}\autoref{fig:J8_Figure_6} are all less than that of{~}\autoref{fig:J8_Figure_4}. Additionally,{~}\autoref{fig:J8_Figure_6} (b) also shows a large area of workspace, but the cables (green lines in{~}\autoref{fig:J8_Figure_2}) between the moving and fixed pulleys may collide with the platform and other cables (red lines in{~}\autoref{fig:J8_Figure_2}).{~}\autoref{fig:J8_Figure_6} (c) shows that increasing the number of fixed pulleys does not contribute to the workspace maximum. In short, targeting the workspace maximum, the configuration shown in{~}\autoref{fig:J8_Figure_2} (a) is the optimal one (considering the collisions between the platform and cables), and we have 4.34\% increase in the reachable workspace.

\begin{figure}[t]\centering
	\includegraphics[width=165mm]{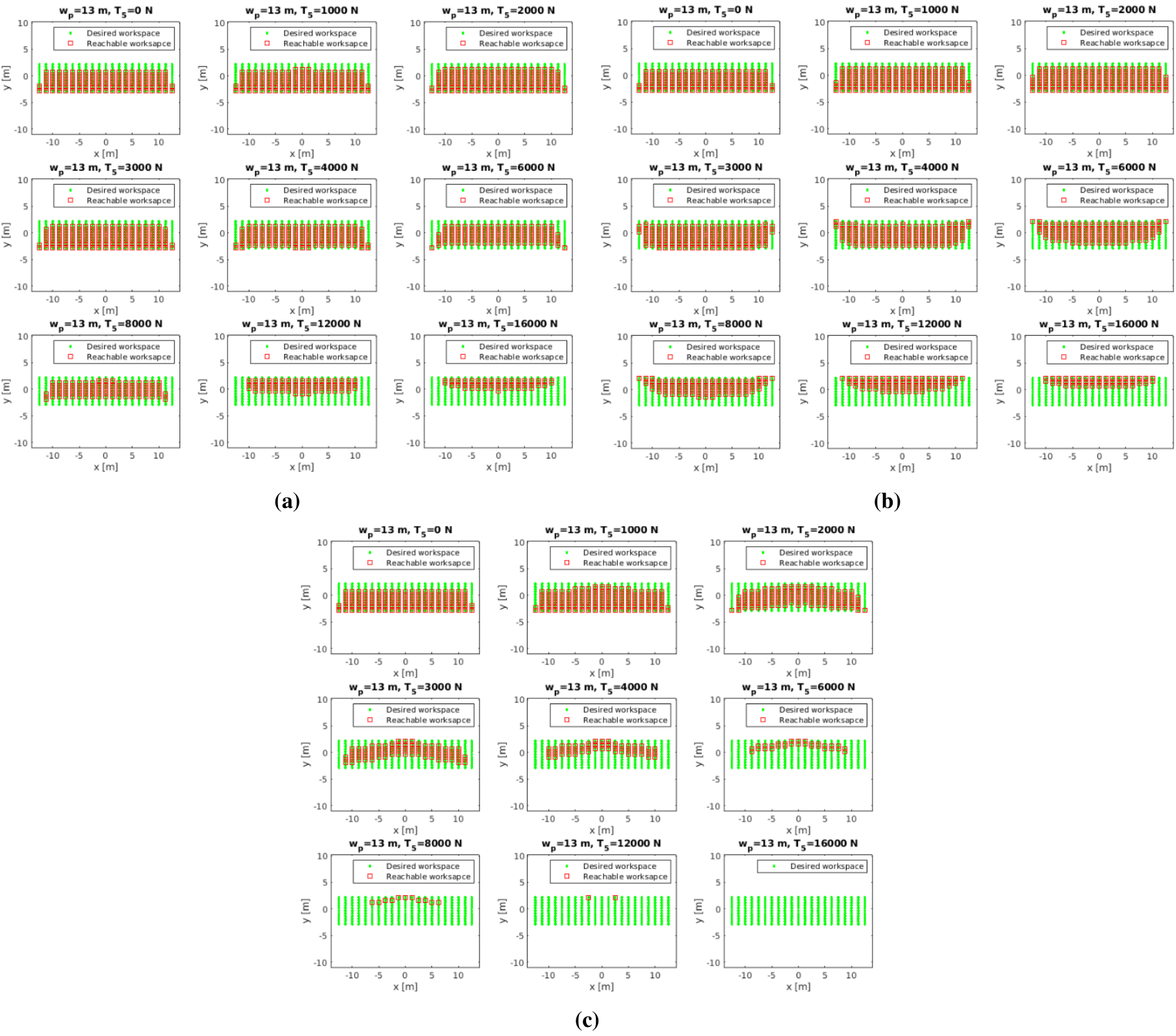}
	\caption{Desired workspace versus reachable workspace of different robot configurations. (a) Responses of the configuration{~}\autoref{fig:J8_Figure_2} (b). (b) Responses of the configuration{~}\autoref{fig:J8_Figure_2} (c). (c) Responses of the configuration{~}\autoref{fig:J8_Figure_2} (d).}
	\label{fig:J8_Figure_7}
\end{figure}

Moreover, the results of the desired workspace versus the reachable workspace of another three robot configurations are provided in{~}\autoref{fig:J8_Figure_7}, in which{~}\autoref{fig:J8_Figure_7} (a), (b), and (c) correspond to the configurations shown in{~}\autoref{fig:J8_Figure_2} (b), (c), and (d), respectively. Compared to the result shown in{~}\autoref{fig:J8_Figure_5}, we can indicate conclusions as follows: Clearly,{~}\autoref{fig:J8_Figure_7} (a) shows a smaller available workspace. Without considering collisions between the moving platform and cables,{~}\autoref{fig:J8_Figure_7} (b) shows a larger vertical feasible workspace to reach the top position of the desired workspace.{~}\autoref{fig:J8_Figure_7} (c) can enlarge the vertical space with a smaller $T_5$ (e.g., at $T_5$ equals 1000 N), but the corners are not reached. In summary, regarding the results shown in{~}\autoref{fig:J8_Figure_5},{~}\autoref{fig:J8_Figure_6}, and{~}\autoref{fig:J8_Figure_7}, we can conclude that the configuration shown in{~}\autoref{fig:J8_Figure_2} (a) is the optimal configuration (aim to maximize the workspace) if collisions are considered; if the collisions are not considered or can be avoided through mechanism design, the configuration shown in{~}\autoref{fig:J8_Figure_6} (b) is the optimal one.

\subsection{Discussion: Active Control of $T_5$} \label{subsec:J8_DiscussionActiveCtrlT5}
\begin{figure}[!t]\centering
	\includegraphics[width=163mm]{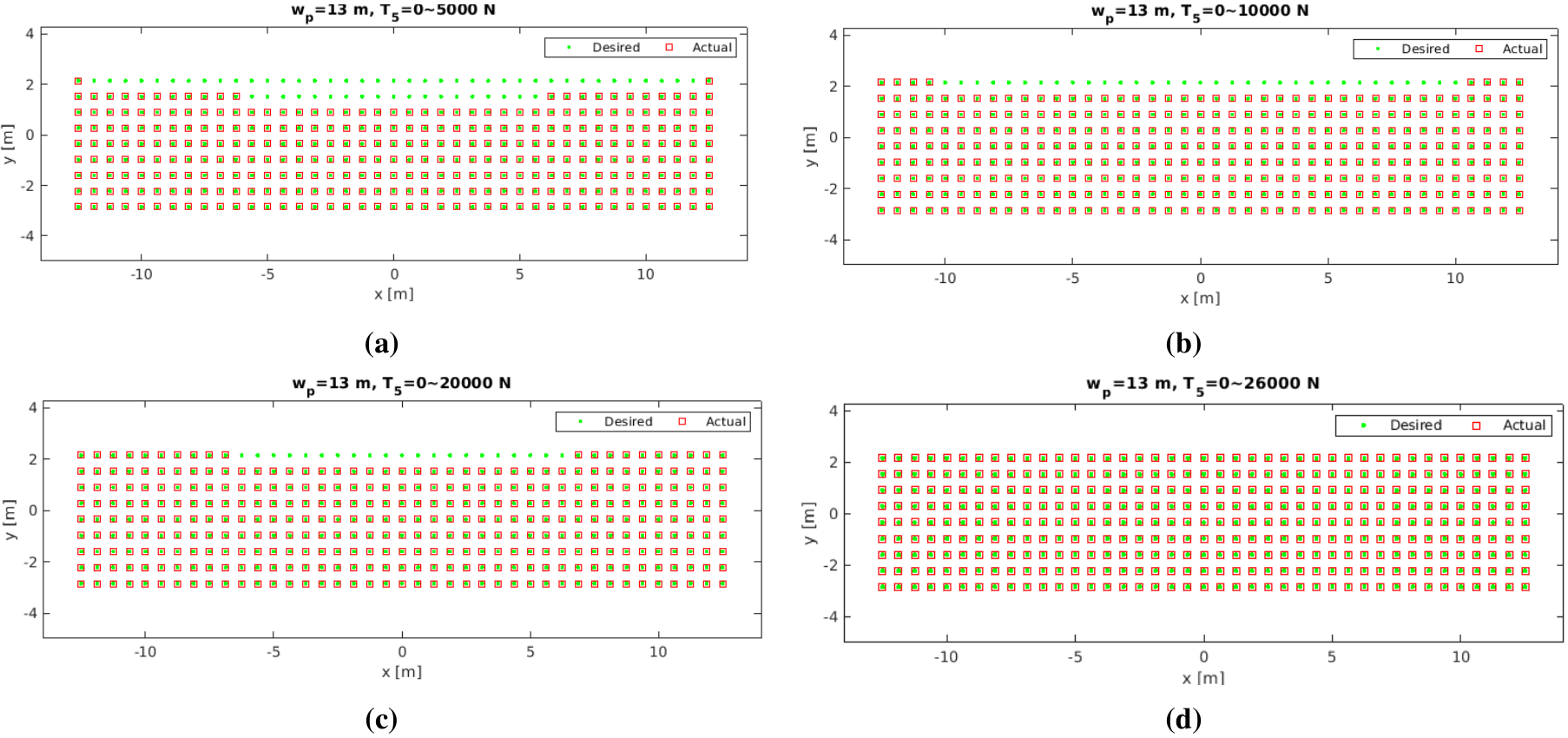}
	\caption{Desired workspace versus reachable workspace with the active control of $T_5$. (a) $T_5$ = 0-5000 N. (b) $T_5$ = 0-10000 N. (c) $T_5$ = 0-20000 N. (d) $T_5$ = 0-26000 N.}
	\label{fig:J8_Figure_8}
\end{figure}

In~\autoref{subsec:J8_ScenarioChooseT5}, we considered $T_5$ as a parameter for workspace improvement and obtained the optimal value. In this scenario, we consider $T_5$ as a variable, i.e., replacing the counterweight by a hydraulic system so that $T_5$ is a control input. In this case, it is interesting to adjust $T_5$ to evaluate whether the reachable workspace can cover the desired workspace. Here, the constraint of $T_{5\max}$ (16000 N) is not is not included to show the maximum coverage of workspace. Using the obtained $w_p$ in~\autoref{subsec:J8_ScenarioChooseWp} and the configuration in{~}\autoref{fig:J8_Figure_2} (a), we conduct this scenario. The results of workspace versus $T_5$ are provided in{~}\autoref{fig:J8_Figure_8}, in which{~}\autoref{fig:J8_Figure_8} (a)-(d) show $T_5$ is equal to 0-5000 N, 0-10000 N, 0-20000 N, and 0-26000 N, respectively. Clearly, the broader region of $T_5$ results in a larger reachable workspace, and the desired workspace can be covered entirely (see{~}\autoref{fig:J8_Figure_8} (d)).

\section{Conclusions and Future Work} \label{sec:J8_Conclusions}
In this paper, we introduced workspace analysis and optimal design of cable-driven parallel robots though auxiliary counterbances. Based on the proposed workspace analysis algorithm, different robot configurations and numerical results were developed for parameters optimization and reachable workspace was increased. Using the proposed approach we had 4.34\% increase in the reachable workspace. Numerical results in different aspects also suggested the effectiveness of the aforementioned approach. In the further, we plan to use the optimal parameters for real system design.

\section*{Acknowledgment}
The authors would like to knowledge the financial support of the Natural Sciences and Engineering Research Council of Canada (NSERC).

\bigskip
\bibliography{J8_Bibliography}

\vfill
\end{document}